\documentclass[a4paper]{article}

\usepackage{INTERSPEECH2019}

\title{Identifying Personality Traits Using Overlap Dynamics in Multiparty Dialogue}
\name{Mingzhi Yu $^1$, Emer Gilmartin $^2$, Diane Litman $^1$}
\address{
  $^1$University of Pittsburgh\\
  $^2$Adapt Centre, Trinity College Dublin}
\email{miy39@pitt.edu, gilmare@tcd.ie, dlitman@pitt.edu}

\begin{document}

\maketitle
\begin{abstract}
  Research on human spoken language has shown that speech plays an important role in identifying speaker personality traits. In this work, we propose an approach for identifying speaker personality traits using overlap dynamics in multiparty spoken dialogues. We first define a set of novel features representing the overlap dynamics of each speaker. We then investigate the impact of speaker personality traits on these  features using ANOVA tests. We find 
  that features of overlap dynamics significantly vary for speakers with different levels of both
  Extraversion and Conscientiousness.
  Finally, we 
  find 
  that classifiers using only overlap dynamics features outperform random guessing in identifying Extraversion and Agreeableness, and that the improvements are statistically significant. 
 
\end{abstract}

\section{Introduction}
Speech has proven to be an important key in identifying speaker personality for the psychological community. Human speech contains cues to personality traits \cite{scherer1979social} and the perception of personality based on speech is highly correlated with the perception of the whole person \cite{ekman1980relative}. In recent decades, a considerable number of studies have obtained encouraging results in recognizing personality traits from human speech data \cite{mohammadi2012automatic, mairesse2007using,gilpin2018perception, pianesi2008multimodal,valente2012annotation}. 

Personality has been described as a constellation of traits or factors derived from the language used to describe them \cite{cattell1943description, norman19672800}. These factors have been operationalized with different models and are frequently measured using self report quizzes such as the Big 5 (OCEAN) \cite{Goldberg1990} and 16PF Questionnaire \cite{cattell2008sixteen}. The Big 5 are five broad dimensions that can be used to describe individual differences in behavior \cite{john1991big}. The Big 5 traits have been described as:

\begin{itemize}
    \item Extraversion (Extrav):  energetic, emotional reactivity, assertive, sociable and talkative.
    \item Agreeableness (Agree): compassionate, cooperative and friendly.
    \item Conscientiousness (Consc): organized, dependable, self-discipline and goal-oriented.
    \item Neuroticism (Neuro):  emotional, anxious and vulnerability.
    \item Openness (Open): intellectual, curious and creative.
\end{itemize}

Personality trait prediction from speech data has been performed on corpora where speakers have completed self-report personality tests at the time of recording - a process known as Automatic Personality Recognition (APR). In the absence of such self-reports, researchers have also used post hoc personality trait descriptions, made by judges reviewing audio or video recordings, to automatically predict personality traits - a process known as Automatic Personality Perception (APP) \cite{schuller2015survey}. A number of studies have used the APR \cite{mairesse2007using, ivanov2011recognition} and APP \cite{mohammadi2010voice, valente2012annotation, gilpin2018perception} paradigms. Mairesse et al. \cite{mairesse2007using} compared the model performance of APP and APR over the same data, and found recognition of observed personality (as used in APP) was more successful than recognition of self-assessed personality (as used in APR), particularly in conversational data. They suggested that the judges observing the personality in the APP task may have used features similar to those ultimately used by the model to make their decisions, and thus were recognizing external cues interpreted as manifestations of personality rather than internal personality as measured in self-assessment. In our work, we use data from self-report Big 5 tests, and are thus recognizing internal or `identity' personality, rather than external or `reputation' personality.

The data used in personality trait identification has included recordings of  single speakers \cite{gilpin2018perception, mohammadi2010voice, mairesse2007using} and spoken dialogue -- both dyadic \cite{ivanov2011recognition} and multiparty \cite{pianesi2008multimodal, valente2012annotation}. Mohammadi et al. \cite{mohammadi2012automatic} and Gilpin et al. \cite{gilpin2018perception} used French broadcast news speech clips to perform a single speaker APP recognition experiment. The corpus contained 640 clips of 10 seconds or less representing 322 speakers, which were annotated with personality traits by a team of 11 judges. Ivanov et al. \cite{ivanov2011recognition} used the PersIA corpus of 2 hours and 14 minutes of simulated tourist call center dialogues (119 calls and 24 identities) to perform an APR experiment. For multiparty dialogue, Pianesi et al. performed an APR experiment using the Mission Survival 2 corpus (over 6 hours) of multiparty groups performing a ranking task, comprising 12 groups of 4 participants each \cite{pianesi2008multimodal}. Valente et al. used a subset of the AMI meetings corpus \cite{carletta2005ami} comprising 12 minute extracts from 32 meetings covering a total of 128 speakers \cite{valente2012annotation}. 

The identification of personality traits from recordings has tended to use features drawn from individual speakers, either speaking alone or in groups. Lexical \cite{mairesse2007using,valente2012annotation}, prosodic  \cite{mohammadi2012automatic,pianesi2008multimodal} and acoustic features \cite{gilpin2018perception,eyben2010opensmile} have been used to predict the personality trait of speakers. Valente et al. \cite{valente2012annotation} included features related to interlocutor interactions when predicting personality traits in multiparty dialogue. Apart from this, few computational studies have attempted to use speaker interactions such as speech overlaps and interruptions as predictive features in identifying speaker personality traits in spoken dialogue. We investigate the effect of interlocutor traits on personality recognition with features extracted from interspeaker activity.

In the psychology literature, simple `on-off' patterns of speech and silence and the resulting pauses, gaps and overlaps have been linked to personality traits at group and individual levels.  For example, at the group level, members of less intelligent groups interrupted more frequently than members in more intelligent groups, while members of less neurotic groups interrupted more than those in more neurotic groups \cite{rim1977personality}. 
At the individual level, interruption rate is negatively correlated to  social anxiety and speech anxiety, and positively correlated to the confidence of a speaker \cite{natale1979vocal}. Individuals who appear less sociable and more assertive tend to interrupt more than those who are not \cite{robinson1989effects}.
Later studies also suggest that interruptions are affected by many variables including their interlocutors' personality traits \cite{Goldberg1990}. Feldstein et al. \cite{feldstein1974personality} found that the personality of both conversational participants influenced the frequency of onset of simultaneous speech (SS). According to him, 
``the extent to which an individual initiates simultaneous speech in a conversation is, indeed, influenced by aspects of his own personality. But it is also influenced ... by personality characteristics of his conversational partner.''
Another interpretation of this claim is that the overall overlap dynamics of a speaker may vary depending on their interlocutors' traits. 

In this study we propose a novel method to represent the overlap dynamics of a speaker while taking account of their interlocutors' traits. We use an APR approach to identify the Big 5 using overlap dynamics in a corpus of multiparty dialogue, containing 213 identities and 62 conversations, where each conversation is approximately 30 minutes long. To our knowledge, this is the first APR work using such overlap dynamics to predict a speaker's Big 5 traits. Although \cite{valente2012annotation} also used speaker overlap and interruption, e.g. speaker interruption frequency, our features of overlap dynamics are based on interlocutors' traits. We use a larger corpus and more speakers than existing work using overlap information in personality trait recognition in multiparty dialogue \cite{valente2012annotation}. We find overlap dynamics vary significantly for speakers with different levels of both Extraversion and Conscientiousness. Classifiers using only overlap dynamics features outperform random guessing in identifying Extraversion and Agreeableness, and the performance improvement is statistically significant.

\section{Teams Corpus}
The Teams Corpus\footnote{https://sites.google.com/site/teamentrainmentstudy/corpus} \cite{litman2016teams} comprises 47 hours of multiparty interaction from 62 teams (35 three-person and 27 four-person teams) playing the collaborative board
game Forbidden Island. This game requires cooperation and verbal communication among the players to win as a group. The team players were 213 native speakers of American English (79M/134F) aged 18 years or older. Each player was assigned to a single team, and each team played two rounds (Game 1 and Game 2) of the game in a single session. The work presented in this paper uses only Game 1 data, where each game is approximately 30 minutes long.
The corpus audio recordings were manually segmented into interpausal units (IPUs), which will serve as the basic unit of analysis in this work,
using a pause length (i.e., silence) threshold of 200 milliseconds.

\begin{figure}[]
  \centering
  \includegraphics[width=0.9\columnwidth, height = 7.5cm]{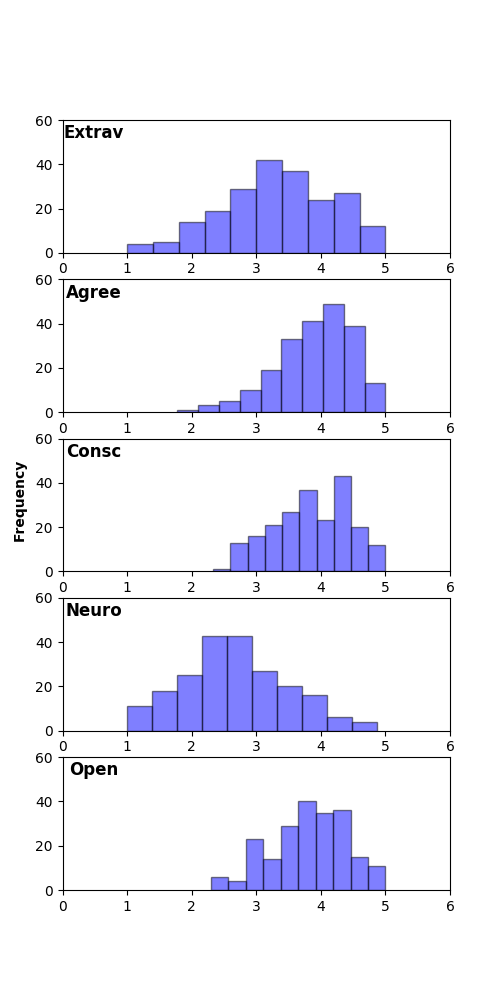}
  \caption{Big 5 score histograms for the Teams corpus.}
  \label{fig:score_histogram}
\end{figure}

Participants took a pre-game survey which included a set of self-report items designed to measure the Big 5 personality traits.  The 44-item Big 5 Inventory (BFI) \cite{john1991big} was used to score participants for each of the Big 5 traits on a scale from 1 (lowest) to 5 (highest).
   Figure \ref{fig:score_histogram} shows the score histograms for the individuals in the Teams corpus, for each trait.   

We further processed the speech files to facilitate the analyses in this paper. All IPUs of less than 500 ms duration were removed, following the methodology used in \cite{kurtic2010resources} for work on overlap in multiparty interaction.  We used Praat to create `floor state' annotations to label speech, silence, and overlap activity for all participants in single labels.  An alphanumeric code for each interval recorded who was speaking during the interval, or labeled intervals of global silence (where nobody was speaking). For example, the label \textbf{aSbS} denotes that speakers \textbf{a} and \textbf{b} are speaking in overlap, while \textbf{cS} indicates that \textbf{c} is speaking alone, and  \textbf{GX} denotes global silence. Any speakers not mentioned in labels are silent for the interval described. These labels could then be used with regex to generate annotations of overlap dynamics.

\section{Trait Measures and Overlap Features}

\subsection{Personality Trait Measures}
\label{sec:measures}
We grouped the speakers in the Teams corpus in two ways, based on their Big 5 personality trait scores. First, for a particular trait, speakers whose scores were greater than or equal to the median were considered to be those who tended to possess this trait. This grouping was used to create the overlap dynamic features described in Section~\ref{sec:features}.
Second, for a particular trait, we automatically labeled each speaker with one of three labels: \textit{Low} ($s < m - 0.5$), \textit{Moderate} ($m - 0.5 <= s <= m + 0.5$) and \textit{High} ($s > m + 0.5$), where {\it s} was the speaker's score  and {\it m} was the corpus median. Here we chose 0.5 as a threshold based on the previous work of \cite{valente2012annotation}.
These labels were used as the dependent measures for the analyses in Section~\ref{sec:analysis}.
Table \ref{tab:subgroup stats} shows the number of speakers in each subgroup. Note that for each trait, the group labeled with \textit{Moderate} has the largest population based on scores. Because speakers at the median were considered to possess the trait, the number of speakers who possess a trait is slightly greater than the number who do not possess a trait.

\begin{table}[]
\centering
\caption{The number of speakers in each subgroup.}
\resizebox{0.9\columnwidth}{!}{
\begin{tabular}{|l|lll||ll|}
\hline
       & Scores &          &      & Possessing &             \\ \hline
       & Low     & Moderate & High & True    & False  \\ \hline
Extrav & 56      & 103      & 54   & 109        & 104         \\ \hline
Agree  & 46      & 133      & 34   & 111        & 102         \\ \hline
Consc  & 40      & 113      & 60   & 118        & 95          \\ \hline
Neuro  & 44      & 118      & 51   & 116        & 97          \\ \hline
Open   & 47      & 140      & 26   & 115        & 98          \\ \hline       
\end{tabular}}
\label{tab:subgroup stats}
\end{table}

\subsection{Overlap Dynamics Features}
\label{sec:features}
\begin{figure}[h]
  \centering
\includegraphics[width=0.95\columnwidth]{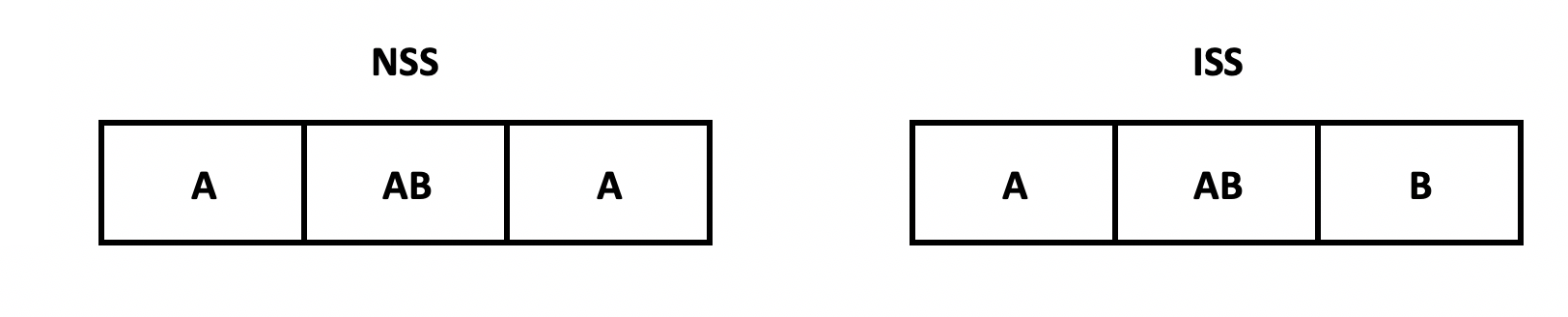}
  \caption{NSS and ISS illustration based on Feldstein et al. \cite{feldstein1974personality}. Left: non-interruptive simultaneous speech (NSS). Right: interruptive simultaneous speech (ISS).
  }
  \label{fig:speech_production}
\end{figure}

We directly adopted two categories of simultaneous speech from \cite{feldstein1974personality} to represent overlap dynamics. For a pair of speakers, A and B, {\it non-interruptive simultaneous speech} (NSS) occurs when B starts speaking while A is already speaking and stops while A continues. {\it Interruptive simultaneous speech} (ISS) occurs when B starts speaking while A is already speaking, and continues speaking after A has stopped speaking. Figure \ref{fig:speech_production} illustrates these two situations. It can be seen that there is no turn change in NSS while the turn passes from A to B in ISS.

In \cite{feldstein1974personality}, one of the interesting findings was that the frequency of simultaneous speech was influenced not only by the speaker's personality characteristics, but also by the partner's personality characteristics. For example, they found relaxed individuals were more likely to have more simultaneous speech than those who were tense. Meanwhile, if the conversational partner was talkative and attentive, it was more likely that more simultaneous speech would occur in the conversation regardless of the speaker's personality. To our knowledge, this finding has not been previously exploited in later computational work. Thus we created a novel set of ISS and NSS features that utilize this finding by taking the personality of the conversational partner into account.
For each speaker, these features represented the average frequency of ISS and NSS between this speaker and conversational partners with particular traits. For speaker {\it i}, the ISS and NSS feature of a trait (i.e.,  one of the Big 5), was defined in Equations \ref{eq:ISS} and \ref{eq:NSS}. Here {\it n} represents the set of conversational partners that possess this trait and {\it j} represents a single speaker. $ISS_{ij}$ and $NSS_{ij}$ represent the frequency of ISS and NSS in the conversation. For example, \textit{Extrav ISS} will denote the average occurrence of ISS between a speaker and speakers who possess Extraversion in the conversation. If speaker i has 5, 10, 12 ISS with speaker p, q and j correspondingly in a conversation and speaker p and q are both Extraversion, then \textit{Extrav ISS} is $(5 + 10)/2 = 7.5$.
\begin{equation}
    trait\ ISS = \frac{(\sum_{j} ISS_{ij})}{|n|}, j \in n 
\label{eq:ISS}
\end{equation}

\begin{equation}
    trait\ NSS = \frac{(\sum_{j} NSS_{ij})}{|n|}, j \in n \\
\label{eq:NSS}
\end{equation}

As speakers were randomly assigned to teams in the corpus,  teams did not always include a speaker  possessing a particular trait. This leads to missing {\it trait ISS} and {\it trait NSS}
values for speakers who can never interact with individuals possessing specific traits. Our handling of missing feature values is discussed in Section~\ref{sec:analysis}.

The ISS and NSS features capture the overlap dynamics of a speaker based on dyadic aspects of team interactions. In multi-party conversation, simultaneous speech also occurs across more than 2 speakers. Therefore, we also created features to represent overlap involving more than two speakers. Here we were particularly interested in overlaps involving 2 speakers and 3 or more speakers since our corpus consisted of 3-party and 4-party conversations. We amalgamated  3 and 4 speaker overlap as  4-speaker overlap was extremely rare in the data -- accounting for less than 0.2\% of total overlap in the corpus. \textit{2 spks overlap} denotes the number of overlaps involving 2 speakers (including the speaker). \textit{3+ spks overlap} denotes the number of overlaps involving at least 3 speakers (including the speaker).

\section{Data Analysis and Model Building}
\label{sec:analysis}

\subsection{Differences in Overlap Dynamics by Trait Strength}
To investigate whether the values of our overlap dynamics features differ significantly across subjects in the \textit{Low}, \textit{Moderate} and \textit{High} trait groups introduced in Section~\ref{sec:measures}, we performed a one-way ANOVA test for each of the Big 5 traits. 
For each analysis, we eliminated missing values. The significant results are shown in Table \ref{tab:ANOVA}. Overlap dynamics were found to significantly vary with Extraversion and Conscientiousness for \textit{Low}, \textit{Moderate} and \textit{High}. For Extraversion, \textit{High} tends to be involved in more 2 speaker and 3+ speaker overlaps than other labeled groups. Also, \textit{High} of Extraversion generally tends to have more ISS with other traits and NSS with Openness. For Conscientiousness, \textit{Low} tends to have more NSS with Agreeableness. Overall, Extraversion emerges as the trait that shows a diverse pattern in different levels. Extroverts overlap with other speakers more frequently than introverts in the conversation. Except for Agreeableness, the interruptive activities occurred frequently between extroverts and other speakers. Compared to introverts,  extroverts generally tend to overlap with other speakers regardless of their personalities. These findings have matched some characteristics of extraversion such as being talkative and energetic, which were mentioned in related literature \cite{norman19672800}. Being low in Conscientiousness can be characterized as being careless and inefficient. The ANOVA result suggests that careless individuals tend to have more non-interruptive overlap with friendly and cooperative persons compared to those who are organized and self-disciplined. 

\begin{table}[]
    \centering
     \caption{One-way ANOVA significant results. * if p \textless 0.05 and ** if p \textless 0.01. 
     M: Moderate, H: High, L: Low. Sample size shows the total number of samples used in ANOVA after eliminating missing data.}
    \resizebox{0.95\columnwidth}{!}{
    \begin{tabular}{|l|l|l|ccc|c|}
    \hline
       & Significant Features & Post hoc                     &       & Mean  &       & Sample Size \\ \hline
       &                      &                              & L     & M     & H     &             \\ \hline
       & 2 spks overlap**       & M \textless H, L \textless H & 182.4 & 197.2 & 243.2 & 213         \\
       & 3 spks overlap*       & M \textless H                & 48.8  & 47.3  & 65.0  & 213         \\
Extrav & Extrav ISS*           & L \textless H, M \textless H & 65.1  & 67.0  & 84.6  & 180         \\
       & Consc ISS*            & L \textless H                & 59.0  & 66.9  & 78.8  & 181         \\
       & Neuro ISS*            & M \textless H                & 63.4  & 64.7  & 80.3  & 182         \\
       & Open ISS*             & M \textless H                & 60.4  & 60.9  & 77.8  & 182         \\
       & Open NSS*             & L \textless H, M \textless H & 23.4  & 23.2  & 30.5  & 182         \\ \hline
Consc  & Agree NSS*            & M \textless L                & 29.3  & 21.7  & 25.9  & 174  
     \\\hline
\end{tabular}}
   
    \label{tab:ANOVA}
\end{table}

\subsection{Predicting Personality Traits using Overlap Dynamics}

Moving from ANOVA analyses to prediction, we constructed a separate Naive Bayes classifier\footnote{We also experimented with SVM and Decision Trees. Here we report Naive Bayes as it achieved the best performance using our data.}  for each of the Big 5, to predict one of the following trait labels from the overlap dynamics features: \textit{Low}, \textit{Moderate} and \textit{High}. To handle  samples with missing feature values, we evolved the imputation strategy in data splitting with K nearest neighbors. We took three steps to partition the dataset into 70\% for training and 30\% for testing. We first divided the data set into two subsets, A (containing only samples with no missing values) and B (the rest of the samples). We randomly drew from subset A a number of samples corresponding to 30\% of the size of the entire dataset. This formed our test set. We then added the remainder of subset A to subset B to form the training set.

We imputed the training set using k-Nearest Neighbour (KNN). To alleviate the class imbalance caused by sampling from different populations, we created 10 train/test splits with the same techniques. The average performance of classifiers over the 10 splits was reported in the evaluation. Also, the performance was compared with a baseline classifier that predicts labels randomly. 

The evaluation results are shown in Table \ref{tab:predict 3 labels}. We compared the performance of the classifier and the baseline over the 10 splits by t-test. 
The classifier for Agreeableness outperformed the baseline, and the improvements in recall and F1 are statistically significant. The Extraversion classifier also outperformed the baseline but only Recall showed significant improvement. Precision for Conscientiousness, and Recall and F1 for Openness were slightly improved but not significantly. Overall, the performance of the classifiers did not appear to be adequately strong for all the Big 5.

\begin{table}[]
    \centering
     \caption{Predicting \textit{Low}, \textit{Moderate} and \textit{High}.  Evaluation metrics: macro averaged Precision (P), Recall (R) and F1. * if p \textless 0.05, ** if p \textless 0.01. Improved performance compared to random is shown in boldface.}
    \resizebox{1\columnwidth}{!}{
        \begin{tabular}{|l|ccc|ccc|}
        \hline
       &                & Naive Bayes    &                &      & Random Guessing &      \\\hline
       & P              & R              & F1             & P    & R               & F1   \\\hline
Extrav & \textbf{0.32}           & \textbf{0.38**}           & \textbf{0.31}           & 0.31 & 0.30            & 0.30 \\\hline
Agree  & \textbf{0.37} & \textbf{0.38*} & \textbf{0.36**} & 0.32 & 0.31            & 0.28 \\\hline
Consc  & \textbf{0.37}           & 0.32           & 0.30           & 0.34 & 0.33            & 0.31 \\\hline
Neuro  & 0.19           & 0.24           & 0.21           & 0.34 & 0.33            & 0.32 \\\hline
Open   & 0.26           & \textbf{0.35}           & \textbf{0.29}           & 0.33 & 0.31            & 0.28\\\hline
\end{tabular}}
   
    \label{tab:predict 3 labels}
\end{table}

We noticed the evaluation of \textit{Low} and \textit{High} showed promising performance depending on the traits. For example, for \textit{High} in Extraversion, Precision (0.24), Recall (0.47) and F1 (0.31) were all statistically significantly higher than  Precision (0.14), Recall (0.24) and F1 (0.17) for the baseline.  To further investigate how the features predict \textit{Low} and \textit{High}, we experimented with only these two labels, following \cite{valente2012annotation}, and saw promising prediction accuracy. After removing all \textit{Moderate} samples, we trained and tested Naive Bayes classifiers. The results are shown in Table \ref{tab:predict 2 labels}. The absolute performance of the classifiers and the baseline were both improved compared to the prediction of \textit{Low}, \textit{Moderate} and \textit{High}. This finding implies that it is easier for classifiers to distinguish between just  \textit{Low} and \textit{High}. With respect to the prediction outcome after utilizing overlap dynamics features, we observed a performance improvement in predicting \textit{Low} and \textit{High} of Extraversion compared to the baseline. The precision, recall and F1 are all improved, and the improvement of F1 is statistically significant. In distinguishing between \textit{Low} and \textit{High} of Agreeableness, the classifier outperformed the baseline significantly. Note that there were some improvements in predicting \textit{Low} and \textit{High} of Conscientiousness and Open but they were not statistically significant. 

In conclusion, only using the overlap dynamics features, Naive Bayes classifiers can achieve up to 29\% relative improvement in F1 compared to the baseline when identifying three labels for a trait (with the highest relative improvement  for Agreeableness). 
The largest relative F1 improvement in identifying just the low and high levels is 14\% for Extraversion, compared to the baseline. 

\begin{table}[]
    \centering
    \caption{Predicting \textit{Low} and \textit{High}. Evaluation metrics: macro averaged Precision (P), Recall (R) and F1. * if p \textless 0.05, ** if p \textless 0.01. Improved performance compared to random is shown in boldface.}
    \resizebox{1\columnwidth}{!}{
    
\begin{tabular}{|l|ccc|ccc|}
\hline
       &                & Naive Bayes    &                &      & Random Guessing &      \\\hline
       & P              & R              & F1             & P    & R               & F1   \\ \hline
Extrav & \textbf{0.57}  & \textbf{0.57}  & \textbf{0.56*} & 0.51 & 0.51            & 0.49 \\\hline
Agree  & \textbf{0.62*} & \textbf{0.63*} & \textbf{0.58*} & 0.54 & 0.54            & 0.53 \\\hline
Consc  & \textbf{0.60}  & 0.53           & 0.51           & 0.56 & 0.56            & 0.55 \\\hline
Neuro  & 0.31           & 0.36           & 0.29           & 0.45 & 0.45            & 0.44 \\\hline
Open   & 0.49          & 0.49           & \textbf{0.47}  & 0.49 & 0.49            & 0.46 \\\hline
\end{tabular}}

\label{tab:predict 2 labels}
\end{table}

\section{Discussion and Future Work}
In this work, we presented an APR approach that uses overlap dynamics in multiparty dialogue. The APR experiments were conducted over a multiparty spoken corpus that contains 213 identities and 62 conversations, where each conversation is approximately 30 minutes long. We introduce a novel representation of overlap dynamics that takes the partner's trait into account. The ANOVA result reveals that the interruption and overlap behaviors may vary between different levels of personality traits. Overall, we found overlap dynamics can be used as indicators of some traits. With respect to APR, models utilizing overlap dynamics features show promising performance. The predictions of Agreeableness and Extraversion show the most significant improvement. Note that in the earlier APP work \cite{valente2012annotation}, the recognition of Agreeableness also shows statistically significant improvement when using their participant interaction features, which only reflect the speaker's overlapping activities but not the overlapping interaction between partners with different traits. It is unclear if this similar finding is a coincidence or  caused by the underlying correlation between Agreeableness and a speaker's overlapping behavior. Future work is needed. 

The current method to determine the possession of traits relies on the median of scores. Since the speaker is also included in the population, it may impact the overall population distribution and result in a potential leak of information in the processing. In the future, we intend to improve the independence of feature extraction. We plan to apply techniques of data splitting such as splitting speakers and their conversational partners. Meanwhile, the current feature extraction of ISS and NSS is based on a supervised dataset, which contains the knowledge of all speakers' personality, including the personality of their conversational partners. In the future, we are interested in approaching the same problem in an unsupervised manner without using the traits of partners. 
The presented study mainly shows that overlap dynamics provided some signal of speaker personality. To further study overlap dynamics, we intend to exam its effectiveness in conjunction with other well-established features in the literature.

\section{Acknowledgements}

This work is supported by the National Science Foundation
under Award \# 1420784. Emer Gilmartin is supported by the ADAPT Centre for Digital Content Technology, which is funded under the SFI Research Centres Programme (Grant 13/RC/2106) and is co-funded under the European Regional Development Fund.

\bibliographystyle{IEEEtran}

\bibliography{main}


\end{document}